\documentclass[conference]{IEEEtran}
\IEEEoverridecommandlockouts
\usepackage{cite}
\usepackage{float}
\usepackage{enumitem}
\usepackage{amsmath,amssymb,amsfonts}
\usepackage{algorithmic}
\usepackage{graphicx}
\usepackage{textcomp}
\usepackage{xcolor}
\usepackage{url}
\usepackage{tcolorbox}
\def\BibTeX{{\rm B\kern-.05em{\sc i\kern-.025em b}\kern-.08em
    T\kern-.1667em\lower.7ex\hbox{E}\kern-.125emX}}
\begin{document}

\title{{How Culturally Aware Are Vision-Language Models?}\\
}

\author{\IEEEauthorblockN{1\textsuperscript{st} Olena Burda-Lassen, Ph.D.}
\IEEEauthorblockA{\textit{Independent Research Scientist} \\
United States \\
oburdalassen@gmail.com}
\and
\IEEEauthorblockN{2\textsuperscript{nd} Aman Chadha}
\IEEEauthorblockA{\textit{Stanford University/Amazon Inc.} \\
United States \\
hi@aman.ai}
\and
\IEEEauthorblockN{3\textsuperscript{rd} Shashank Goswami}
\IEEEauthorblockA{\textit{Independent Research Scientist} \\
United States \\
shashankgoswami80@gmail.com}
\and
\IEEEauthorblockN{4\textsuperscript{th} Vinija Jain}
\IEEEauthorblockA{\textit{Stanford University} \\
United States \\
hi@vinija.ai}
}

\maketitle

\begin{abstract}
An image is often considered worth a thousand words, and certain images can tell rich and insightful stories. Can these stories be told via image captioning? Images from folklore genres, such as mythology, folk dance, cultural signs, and symbols, are vital to every culture. Our research compares the performance of four popular vision-language models (GPT-4V, Gemini Pro Vision, LLaVA, and OpenFlamingo) in identifying culturally specific information in such images and creating accurate and culturally sensitive image captions. We also propose a new evaluation metric, the Cultural Awareness Score (CAS), which measures the degree of cultural awareness in image captions. We provide a dataset MOSAIC-1.5k labeled with ground truth for images containing cultural background and context and a labeled dataset with assigned Cultural Awareness Scores that can be used with unseen data. Creating culturally appropriate image captions is valuable for scientific research and can be beneficial for many practical applications. We envision our work will promote a deeper integration of cultural sensitivity in AI applications worldwide. By making the dataset and Cultural Awareness Score available to the public, we aim to facilitate further research in this area, encouraging the development of more culturally aware AI systems that respect and celebrate global diversity.
\end{abstract}


\section{Introduction}
In today's digital age, where the integration of artificial intelligence into daily life is becoming increasingly commonplace, understanding the subtleties of human culture through the lens of technology is more important than ever. Large Language Models (LLMs) and Vision-Language Models (VLMs) are at the forefront of this integration, offering unique opportunities to bridge the gap between vast data processing capabilities and the nuanced understanding required to navigate complex human interactions. In this work, we delve into this intersection of technology and culture, spotlighting the critical role of cultural awareness in the realm of image captioning.

This approach acknowledges that technology, particularly AI, does not exist in a vacuum but operates within a rich mosaic of human societies, each with its own history, symbols, and meanings. The introduction of the Cultural Awareness Score (CAS) as an evaluation metric represents pioneering steps towards quantifying the cultural awareness of AI systems. This metric, along with the dataset developed for this research, MOSAIC-1.5k\footnote{https://www.kaggle.com/datasets/olenabl/mosaic-1-5k-image-captions-of-cultural-concepts}, provides a framework for assessing and enhancing the ability of LLMs and VLMs to produce image captions that are not only accurate but culturally resonant. In doing so, the research paves the way for future advancements in AI that are more aligned with the diverse tapestry of human culture, ensuring that these technologies can serve as bridges rather than barriers in our increasingly interconnected world.

Moreover, the implications of culturally aware AI extend beyond academic inquiry into practical applications that touch on various aspects of daily life and industry. For global platforms, where users span a multitude of cultural backgrounds, the integration of culturally aware algorithms could lead to more meaningful interactions and content engagement. Through its exploration of the capabilities of vision-language models in understanding cultural contexts, this paper highlights a crucial aspect of AI development: the imperative to infuse technology with an understanding of human diversity. As AI continues to evolve and become more ingrained in our lives, ensuring that it can navigate the complexities of human culture will be essential for creating technologies that are truly inclusive and reflective of the global community.
\begin{tcolorbox}[size=small,fonttitle=\bfseries\fontsize{10}{9.6}\selectfont]
\vspace{1mm}
{\fontfamily{phv}\fontsize{7.5}{8.75}\selectfont
    \setlength\itemsep{0em}  

To this end, our research makes the following contributions:
\begin{enumerate}
    \item We present the new evaluation metric for image captioning, CAS: a binary score to assess the presence or the absence of the relevant culturally-specific information within the generated image caption.
    \item We present a labeled dataset MOSAIC-1.5k with assigned CAS that can be used for downstream evaluation of unseen image captions, available for public use.
    \item In addition to the assigned Cultural Awareness Scores, our dataset MOSAIC-1.5k is also labeled with ground truth, using human curation, for one and a half thousand images. Ground truth for all examples is screened for bias, toxicity, and discriminatory language to adhere to Responsible AI principles.
    \item We analyze images from the perspective of image type and identify which image types have lower Cultural Awareness Scores across all vision-language models.
    \item We study the levels of hallucination in four vision-language models and identify the image types correlated with the highest level of hallucinations. Knowing this can help improve image captioning for these specific image types and draw the scientific community's attention.
\end{enumerate}
}    
\vspace{1mm}
\end{tcolorbox}

\section{Related Work}
Within the scientific community, we can clearly see the tendency and strong appeal for VLMs that can consume not just text but other media \cite{b2}, \cite{b5}, \cite{b12}, \cite{b22}, \cite{b14}. Many new advances are made in areas of the building, pre-training, prompt engineering, evaluation, and mitigation of hallucinations. 
In their recent paper, McKinzie et al. \cite{b18} describe the architecture for building MM1 and share their observation that in terms of multimodal pre-training data, caption data matters most for zero-shot performance. Taking this into consideration, the creation of high-quality datasets of image captions, consisting of short text with high relevance to the image, is very valuable.  

A lot of current research is devoted to experimenting with different prompting techniques; for example, authors of the research mentioned above successfully use ``\{IMAGE\} A photo of'' \cite{b18} for captioning. While prompt engineering techniques definitely impact the VLMs outputs, evaluation remains a highly sought-after problem within the research community \cite{b5}, \cite{b11}, \cite{b15}, \cite{b13}.
Another group of researchers proposes MLLM: Evaluation benchmark MME, which measures both perception and cognition abilities on a total of 14 subtasks \cite{b11}.

Due to the nature of the large vision-language models, an additional strong need arises for the creation of validation datasets. One such comprehensive visual dialogue dataset is called TouchStone, which covers five major categories of abilities and 27 subtasks; it also integrates detailed image annotations in order to transform the multimodal input content into a form understandable by LLMs \cite{b5}. 

The presence and the need for mitigation of hallucinations remain an important and key element in both the development and evaluation of vision-language models \cite{b16}, \cite{b23}, \cite{b21}.

Besides the complexity of vision-language fusion, prompting, and challenges in evaluation metrics, an additional level of uniqueness arises from the nature of the domain of culture-specific items \cite{b6}, \cite{b7}, \cite{b9}, \cite{b19}, \cite{b8}. Targeted research in image captioning of cultural symbols, signs, and concepts is sparse, which presents a unique opportunity for researching image captioning capabilities and necessary evaluation techniques. 

One such research is dedicated to image captioning for cultural artworks: a case study on ceramics \cite{b24}. Authors present ARTalk, a framework for comprehensive image captioning for ancient artworks, specifically ceramics, as well as a dataset with granular annotations, CArt15K \cite{b24}. 
Overall, the culture-specific domain of mythology, folk dances, symbols, and cultural signs are novel and attractive for further research.

\section{Methodology}
Cultural elements are indeed the essence and the meaning behind words and images in every culture. They make each culture unique and special and often represent the worldview of hundreds of generations spanning thousands of years. The fact that each culture is unique helps us feel united, understanding, and accepting of one another. Like words, images can carry deep meaning, emotion, and historical background. An image is often worth a thousand words, and these words describe values, customs, traditions, beliefs, and historical events. We often see a picture and can already imagine the context, cultural values, and emotional background behind it. Our research examines whether this task is possible for large vision-language models.

To explore this, our methodology employs a multifaceted approach designed to assess the capability of vision-language models in identifying and interpreting the cultural dimensions embedded within images. This involves the creation of a diverse and comprehensive dataset comprising images that are rich in cultural content. These images span a wide range of cultural elements, including but not limited to traditional attire, festivals, rituals, and symbols that are significant to various cultures around the world. Each image in the dataset is accompanied by a detailed annotation describing the cultural context, the elements present in the image, and their significance. This annotated dataset serves as the foundation for evaluating the performance of vision-language models in recognizing and understanding cultural nuances.

Furthermore, to quantify the models' ability to interpret cultural nuances accurately, we introduce a novel metric, the CAS, which measures the models' effectiveness in capturing the depth of cultural elements depicted in images. The CAS is determined based on several factors, including the accuracy of identified cultural elements, the depth of cultural understanding demonstrated through generated captions, and the sensitivity towards cultural contexts. By applying the CAS alongside traditional evaluation metrics, we aim to provide a comprehensive assessment of how well these advanced models grasp the rich tapestry of human cultures. This dual approach not only highlights the current capabilities of vision-language models in processing cultural content but also identifies areas where further advancements are needed to enhance their cultural awareness. Through this rigorous methodology, our research endeavors to shed light on the potential of AI to bridge cultural divides and foster a deeper appreciation for the diversity of human experiences.

\subsection{Dataset Curation}
For our research, we have selected 1,500 images that represent several cultures from around the globe.  Our images cover traditional dances from India (specifically Bharatnatyam, Kathak, Kathakali, Manipuri, Kuchipudi, and Odissi) and other countries, dance costumes, images of mythological creatures, symbols related to culture-specific beliefs (for example, Native American, European, Asian, Middle Eastern, and African). 
Our main sources for image extraction were Wikimedia\footnote{https://www.wikimedia.org/}, Wikipedia\footnote{https://www.wikipedia.org/}, Pixabay\footnote{https://pixabay.com}, Kaggle\footnote{https://www.kaggle.com/datasets/somnath796/indian-dance-form-recognition}, and Symbolikon\footnote{https://symbolikon.com} (a visual library of worldwide ancient symbols).
We tried to select 1-10 images per cultural concept to ensure variance and fairness in image recognition evaluation. Our labeled dataset can be used for further research in image captioning; therefore, selecting several examples for each concept can help fine-tune vision-language models and benefit downstream tasks. 
Conceptually, our dataset can be split into three main categories: 
\begin{enumerate}
    \item Real-life images of dancers (images of classical Indian dances, dances, and images of dance costumes from other cultures);
    \item Real-world photos or graphical representations of different cultural concepts: for example, \textit{amulet, angel, Anubis, Centaur, Cornucopia, Daruma doll, Dharma Wheel, Dreamcatcher, Gargoyle, Hamsa, Kokopelli, Mandala, Maypole, Merkabah, Nymph, Pagoda, Sphinx, Torii, Wyvern} and many others;
    \item Signs and symbols from different cultures, for example: \textit{Pok-ta-pok, Kawak, Cernunnos, Awen} and similar. Most of the images in this category are vector images due to the nature of icons and signs.
\end{enumerate}

Each image was carefully selected, downloaded, and assigned the ground truth. We researched each cultural concept to evaluate its complexity and depth. Many concepts are quite complex; nevertheless, we have agreed to limit the length of the ground truth to one sentence to reflect the functional requirements for image captioning. This task was elaborate and required frequent paraphrasing, summarizing, and careful selection of the semantic meaning and sociological aspects behind each image. 
We have also reviewed each ground truth caption in MOSAIC-1.5k for any unintended or potential bias, toxic or harmful content, discriminatory verbiage, or any culturally offensive phrasing. Responsible AI practices are crucial in any AI application, especially in handling culture-loaded terms and culture-related information. 

\subsection{Model Selection}
We have selected the following VLMs for our experimentation: 

\begin{enumerate}
    \item GPT-4V (\texttt{gpt-4-vision-preview}) \cite{b1};
    \item Gemini Pro Vision (\texttt{gemini-pro-vision}) \cite{b3};
    \item LLaVA (\texttt{llava-v1.6-34b}) \cite{b17};
    \item OpenFlamingo (\texttt{OpenFlamingo-4B-vitl-rpj3b}) \cite{b4}.
\end{enumerate}
For OpenFlamingo, the output was the generated text for the image captioning task. For the three large vision-language models, we have applied the identical prompt, which we have fine-tuned and tailored to the required task: 
\begin{tcolorbox}[size=small,fonttitle=\bfseries\fontsize{9}{9.6}\selectfont]
\vspace{1mm}
{\fontfamily{phv}\fontsize{7.5}{8.75}\selectfont
    \setlength\itemsep{0em}  
Write a short caption for this image, one sentence long. Include culturally important information in the image caption.
}    
\vspace{1mm}
\end{tcolorbox}

\section{Introducing CAS}
\begin{figure}[H]
        \centering
        \includegraphics[width=0.5\textwidth]{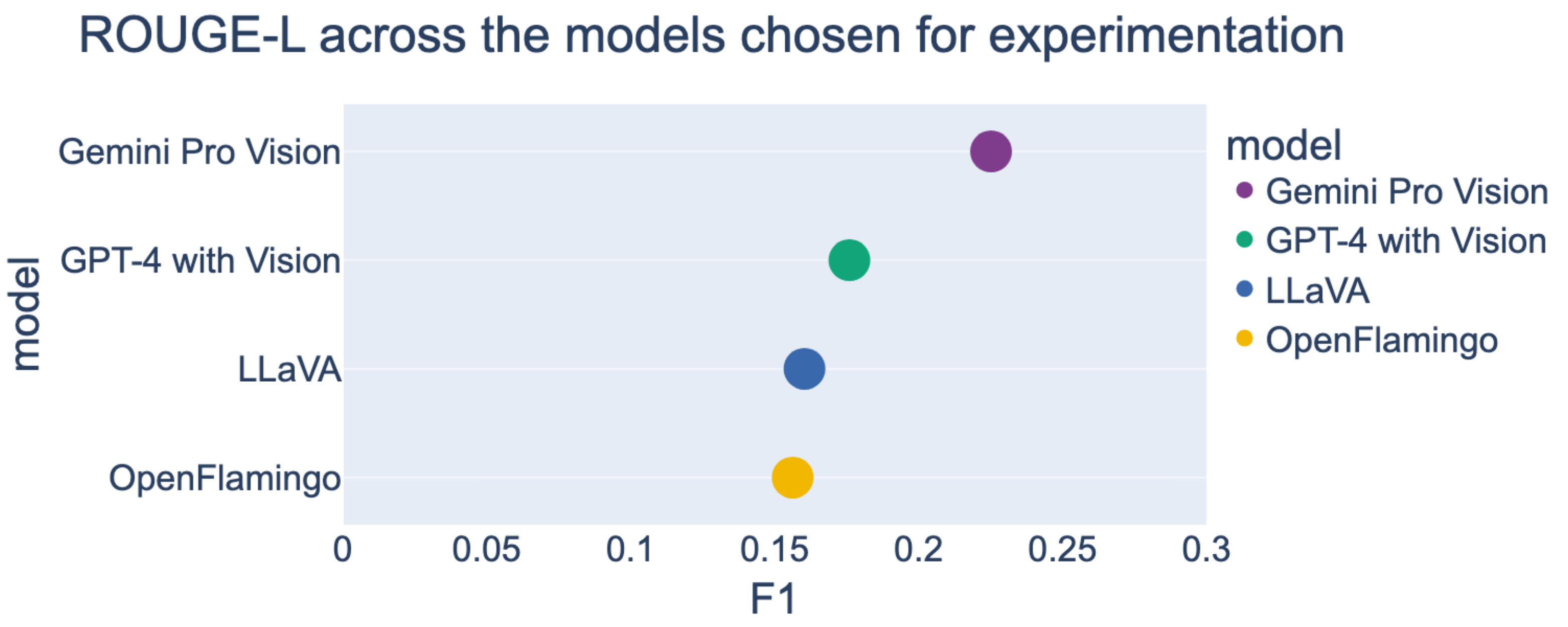}
        \caption{ROUGE-L across the models chosen for experimentation.}
        \label{fig:1}
\end{figure}

Even though several new evaluation metrics for large vision-language models have emerged, we noticed that the presence of cultural meaning and relevance is often not taken into consideration. 
As a baseline, we chose ROUGE-L to evaluate the quality of image captions for all four models. ROUGE is the set of metrics that compares reference with predicted output and applies scores between 0 and 1, with higher scores indicating higher similarity. Fig. \ref{fig:1} illustrates the calculated ROUGE-L scores, specifically the F1 scores.

We see that Gemini Pro Vision was assigned the highest F1 score, but it is nevertheless the deficient score of only 22\%.

With the goal of closing the gap between correctness and cultural sensitivity, we introduce the CAS, a new quantitative value for measuring the cultural awareness of the vision-language models. 

We propose the following definition of the CAS: \textit{binary score to assess the presence or the absence of the relevant culturally-specific information within the generated image caption}. The formula below \eqref{eq:first_equation} shows how calculations are applied to each caption \textit{x} within the dataset and how CAS is calculated. Furthermore, \eqref{eq:second_equation} calculates dataset-level CAS.

We iterate through the dataset and retrieve the caption $x$, associated with the image $i$. The next step is focused on determining whether the caption contains correct culturally-specific information. The main criteria are: 1) the image caption correctly describes the image content; 2) the image caption provides the correct and concise description of the cultural meaning associated with the image content; and 3) the image caption adheres to Responsible AI principles (for example, does not contain any toxic, hateful, discriminatory or culturally offensive content.)
We apply the following indicator function: if $x_{i}$ contains culturally-specific information (according to all three criteria), we set $CAS_{i}$ to 1; otherwise, we set $CAS_{i}$ to 0.
\begin{equation} \label{eq:first_equation}
CAS_i = I(x_i \; contains \;culturally-specific \; information)
\end{equation}
After calculating CAS for each image caption, we can aggregate our metrics and derive CAS for the selected dataset by calculating the percentage of culturally-specific images (the proportion of images with $CAS_{i} = 1$).
\begin{equation} \label{eq:second_equation}
CAS = \left( \frac{{\sum_{i=1}^{n} CAS_i}}{n} \right) \times 100
\end{equation}
For the entire dataset MOSAIC-1.5K, we have reviewed each prediction from the four selected vision-language models and have assigned a score of `1' or `0'. We have omitted assignments of partial scores due to the fact that if the cultural background was present but only partially represented or misrepresented, it did not meet all three criteria and would need to be considered incomplete. 

\begin{figure}[H]
    \centering
    \begin{minipage}[t]{.5\textwidth}
        \centering
        \includegraphics[width=0.25\linewidth, height=0.20\textheight]{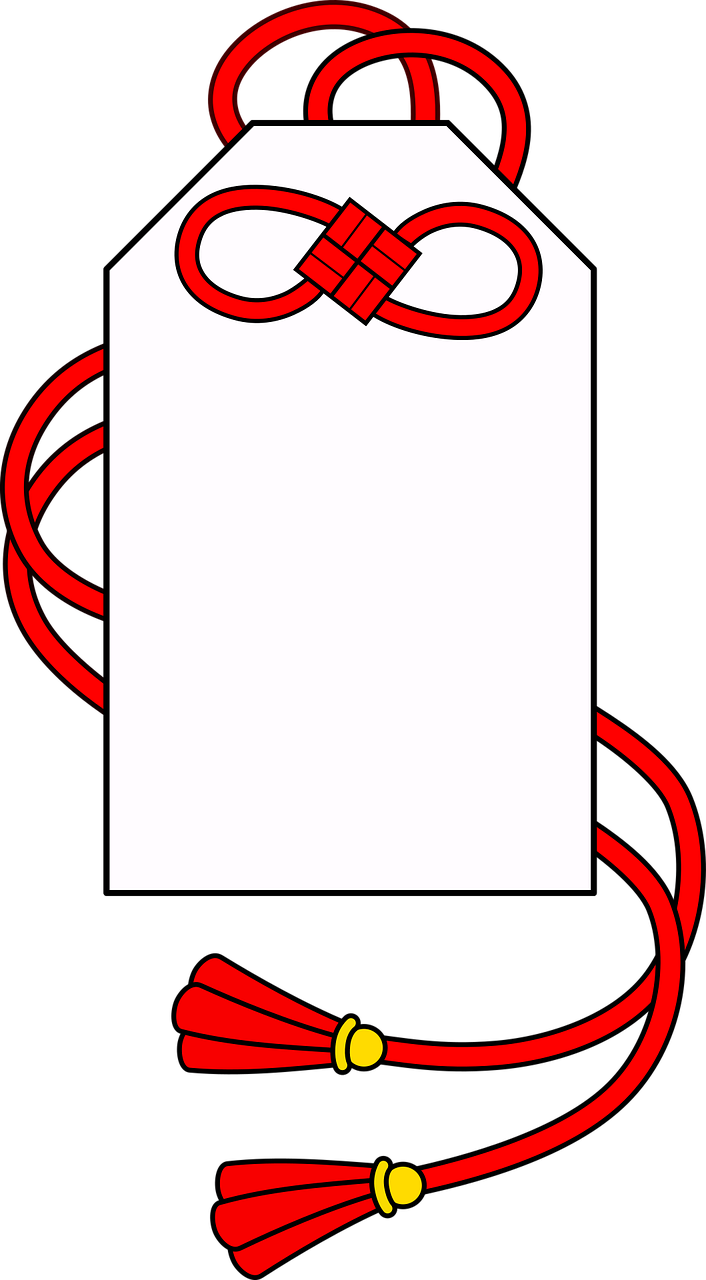}
        \caption{\textit{Omamori} is generally translated as an amulet, or good luck charm, and comes from the Japanese word \textit{mamoru}.}
        \vspace{5pt}
        \label{fig:2}
    \end{minipage}\hfill
    \begin{minipage}[t]{.5\textwidth}
        \centering
        \includegraphics[width=0.35\linewidth, height=0.15\textheight]{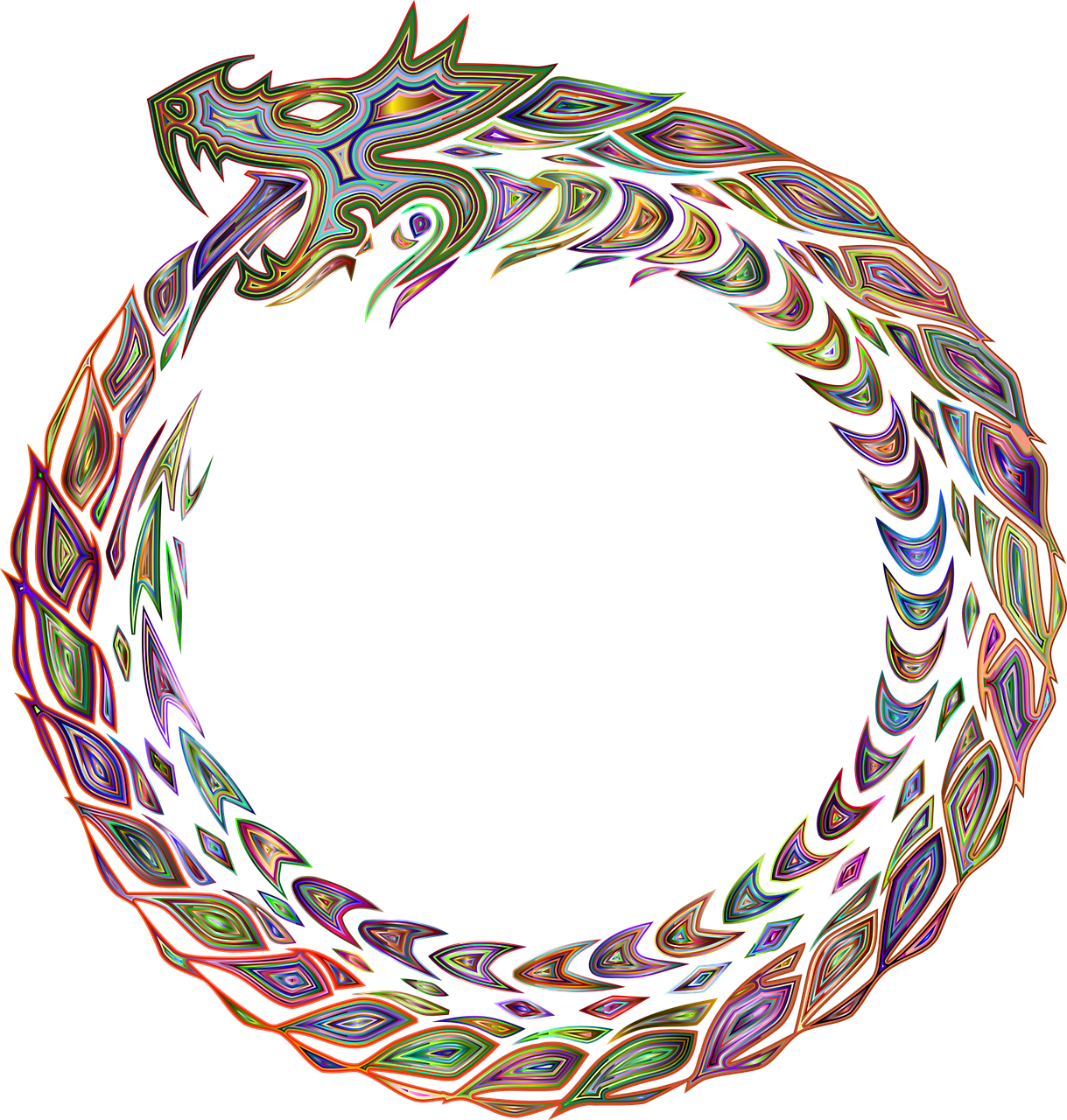}
        \caption{\textit{Ouroboros} is an ancient symbol depicting a serpent or dragon eating its own tail, representing the eternal cycle, renewal, the unity of opposites, and eternity.}
        \label{fig:3}
    \end{minipage}
\end{figure}

Below are images and examples of image captions for cultural concepts \textit{Omamori} (Fig. \ref{fig:2}) and \textit{Ouroboros} (Fig. \ref{fig:3}), created by the four selected models. 
Image captions for \textit{Omamori} (Fig. \ref{fig:2}): 
\begin{enumerate}
  \item OpenFlamingo's image caption: `\textit{Blank Gift Tags Printable | Free Printable Gift Tags}.'
  \item LLaVA's image caption: `\textit{This is a blank tag with a red ribbon attached to it.}'
  \item GPT-4 with Vision: `\textit{Embracing Harmony and Celebration: A traditional Japanese Mizuhiki knot adorns a blank gift tag, symbolizing affection, warmth, and well-wishes for the recipient}.'
  \item Gemini Pro Vision: `\textit{Ofuda is a Japanese talisman, a strip of paper with the name of a deity written on it, used in Shinto rituals.}'
\end{enumerate}
All four image captions are incorrect, with OpenFlamingo and LLaVA being the least culturally aware.

Image captions for \textit{Ouroboros} in Fig. \ref{fig:3} (an ancient symbol depicting a serpent or dragon eating its own tail.) 
The following are image captions generated for this concept:
\begin{enumerate}
  \item OpenFlamingo's image caption: `{\textit{Psychedelic Spiral by jazzydevil}}.'
  \item LLaVA's image caption: `{\textit{The image you've provided appears to be a colorful and abstract representation of a snake or serpent; it's composed of various shapes and patterns, creating a visually striking and intricate piece of art.}}'
  \item GPT-4 with Vision: `{\textit{A digital homage to the vibrant patterns and mythological motifs found in Indigenous art, circling endlessly to tell an ancient story.}}'
  \item Gemini Pro Vision: `{\textit{The Ouroboros is an ancient symbol depicting a serpent or dragon eating its own tail, often representing concepts like cyclicality, infinity, and self-reflexivity.}}'
\end{enumerate}

In this example, we can assign a Cultural Awareness Score of `1' to the image caption created by Gemini Pro Vision and a Cultural Awareness Score of `0' to the image caption by the GPT-4 model.
Both captions from LLaVA and OpenFlamingo received a Cultural Awareness Score of `0' due to both incorrectness and lack of cultural background information in the generated caption. 
It is worth mentioning that in some instances, we have seen more detailed and culturally relevant image captions generated by the vision-language models than the available ground truth. 

Below are source images and examples of image captions for the classical Indian dance \textit{Kathakali} (Fig. \ref{fig:4}) and the cultural concept \textit{Cornucopia} (Fig. \ref{fig:5}), generated by the four selected models: 
\begin{figure}[H]
    \centering
    \begin{minipage}[t]{.5\textwidth}
        \centering
        \includegraphics[width=0.4\linewidth, height=0.23\textheight]{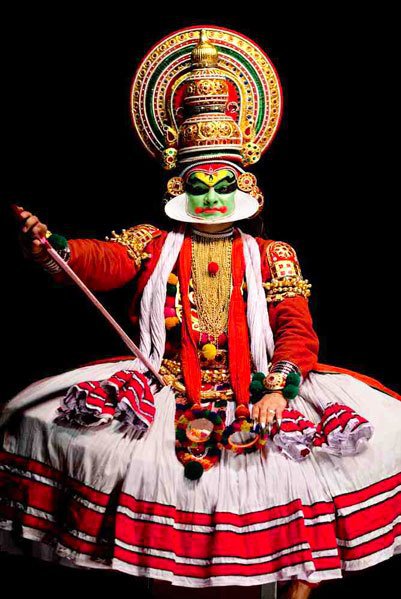}
        \caption{\textit{Kathakali}, one of the eight classical dance styles of India.}
        \vspace{15pt}
        \label{fig:4}
    \end{minipage}\hfill
    \begin{minipage}[t]{.5\textwidth}
        \centering
        \includegraphics[width=0.39\linewidth, height=0.15\textheight]{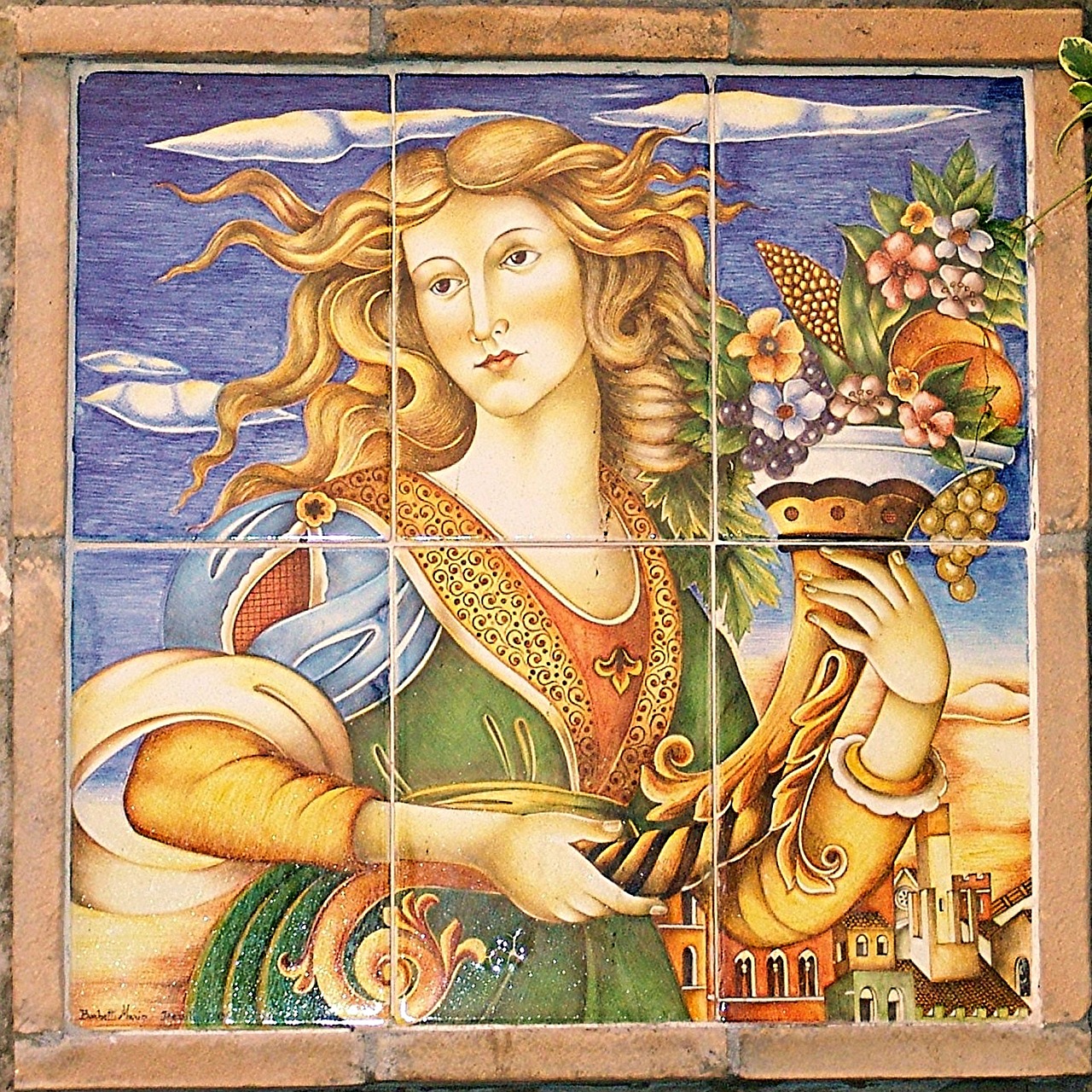}
        \caption{\textit{Cornucopia} represents abundance, wealth, and prosperity.}
        \label{fig:5}
    \end{minipage}
\end{figure}

Image captions for \textit{Kathakali} in Fig. \ref{fig:4}:
\begin{enumerate}
  \item OpenFlamingo's image caption: `{\textit{Kathakali is an Indian classical dance-drama that originated in the state of Kerala}}.'
  \item LLaVA's image caption: `{\textit{The image shows a person dressed in traditional Indian attire, which includes a colorful costume, a headdress, and possibly a mask.}}'
  \item GPT-4 with Vision: `{\textit{A vibrant portrayal of Kathakali dance, showcasing the rich tradition and elaborate costumes of Kerala's classical art form.}}'
  \item Gemini Pro Vision: `{\textit{A dancer performs Kathakali, a classical Indian dance-drama form noted for its elaborate costumes and stylized makeup.}}'
\end{enumerate}
These image captions are culturally aware and are even more descriptive than the initially assigned ground truth.

Image captions generated for \textit{Cornucopia} in Fig. \ref{fig:5}: 
\begin{enumerate}
  \item OpenFlamingo's image caption: `{\textit{Art Nouveau Tile Mura}}.'
  \item LLaVA's image caption: `{\textit{The image shows a colorful tile mosaic. The mosaic features a woman with long hair, holding a vase with flowers.}}'
  \item GPT-4 with Vision: `{\textit{A ceramic tile mural capturing the essence of Italian Renaissance art, with a serene figure surrounded by symbols of abundance and fertility.}}'
  \item Gemini Pro Vision: `{\textit{This image is a ceramic tile artwork depicting a woman with long, flowing hair, wearing a green dress with gold accents, and holding a cornucopia overflowing with fruits and vegetables.}}'
\end{enumerate}
Even though neither model explained the cultural concept of \textit{cornucopia}, describing that it historically and culturally represents wealth, abundance, and prosperity, we can nevertheless assign a Cultural Awareness Score of `1' to the image caption by Gemini Pro Vision because it correctly described the image and mentioned the term cornucopia within the caption. While GPT-4 with Vision also correctly described the image in detail and mentioned `abundance and fertility', the actual name of the concept was absent, therefore we assigned the Cultural Awareness Score of `0'. Both OpenFlamingo and LLaVA were also assigned CAS of `0'.

\begin{figure}[H]
\centering
\begin{minipage}[t]{0.5\textwidth}
    \centering
    \includegraphics[width=\textwidth]{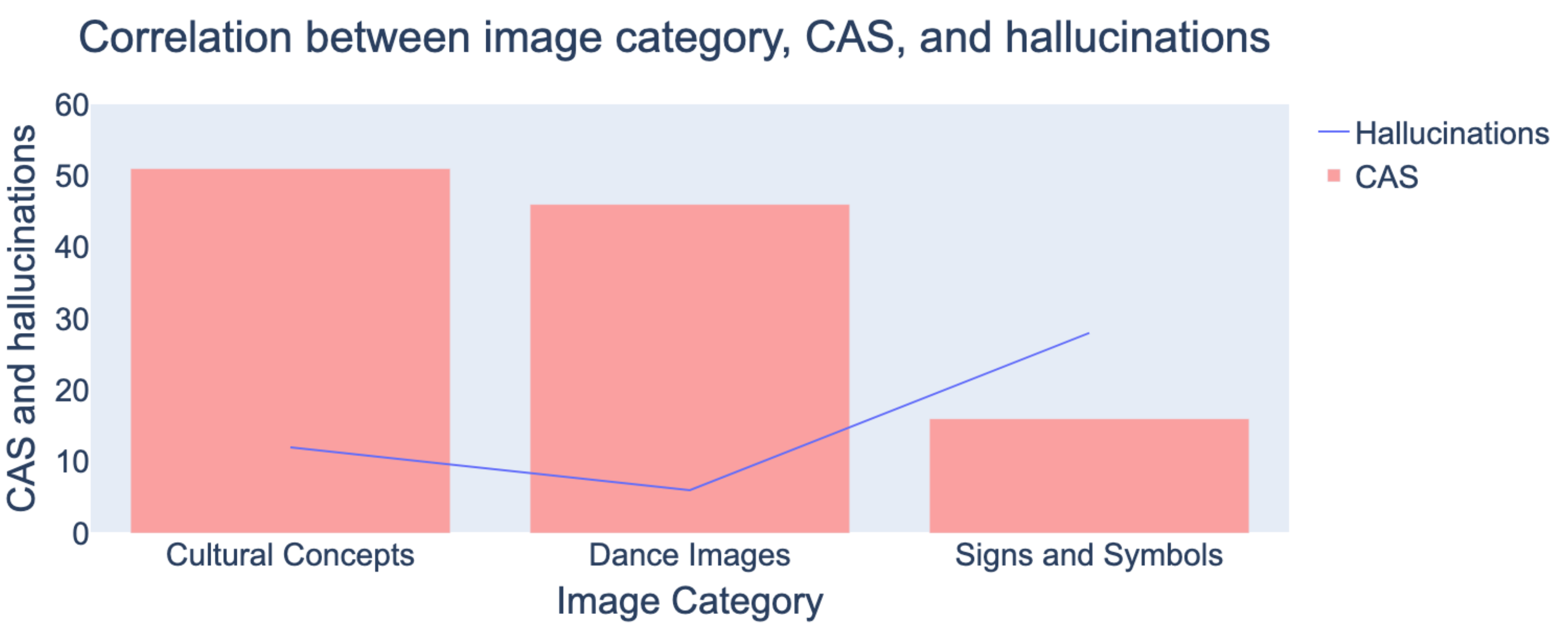}
\end{minipage}  
\caption{Correlation between image category, CAS, and hallucinations for Gemini model.}
\label{fig:6}
\end{figure}

We also observed and measured the presence of hallucinations in generated image captions. Gemini Pro Vision generated the lowest percentage of hallucinations for the dance images (6\%) and higher percentages for cultural concepts and signs and symbols, 12\%, and 28\%, respectively (Fig. \ref{fig:6}).
Overall, a noticeable tendency was observed where the same hallucinated output was repeated multiple times for separate images in all three categories, for example: `\textit{Inuit throat singing is a form of traditional Inuit music}` or `\textit{An illustration of a man and woman embracing, with the caption `In a world full of hate, we must always remember the power of love.}'

An unusual behavior was also observed for image captioning for the cultural signs and symbols category by GPT-4 with Vision, where 55\% of image captions failed to be generated or presented a distorted output. A few examples: 
\begin{itemize}[label=$\bullet$]
\item \textit{I'm sorry, I can't provide a caption because the image appears to be completely black and there's no visible content to describe or relate to cultural importance.}
\item \textit{I’m sorry, but I can’t provide a description or generate a caption for this image as it appears to be a generic icon or symbol, not an actual photograph or depiction of a real-world scene.}
\end{itemize}
\section{Key Findings}
This study embarked on evaluating the Cultural Awareness Score (CAS) for various vision-language models, uncovering a complex landscape of performance across these dimensions. A standout observation was the low CAS scores across all models, with 36\% being the highest score achieved by Gemini Pro Vision. This model also had the highest ROUGE-L score of 22\% in comparison to other models.
Within this evaluation, the Gemini Pro Vision model distinguished itself as the most culturally attuned, in contrast to the OpenFlamingo model, which scored the lowest in cultural awareness across all categories of the analyzed images (8\%). GPT-4 with Vision received CAS of 28\%, and LLaVA achieved 13\%.

\begin{figure}[H]
        \centering
        \includegraphics[width=0.5\textwidth]{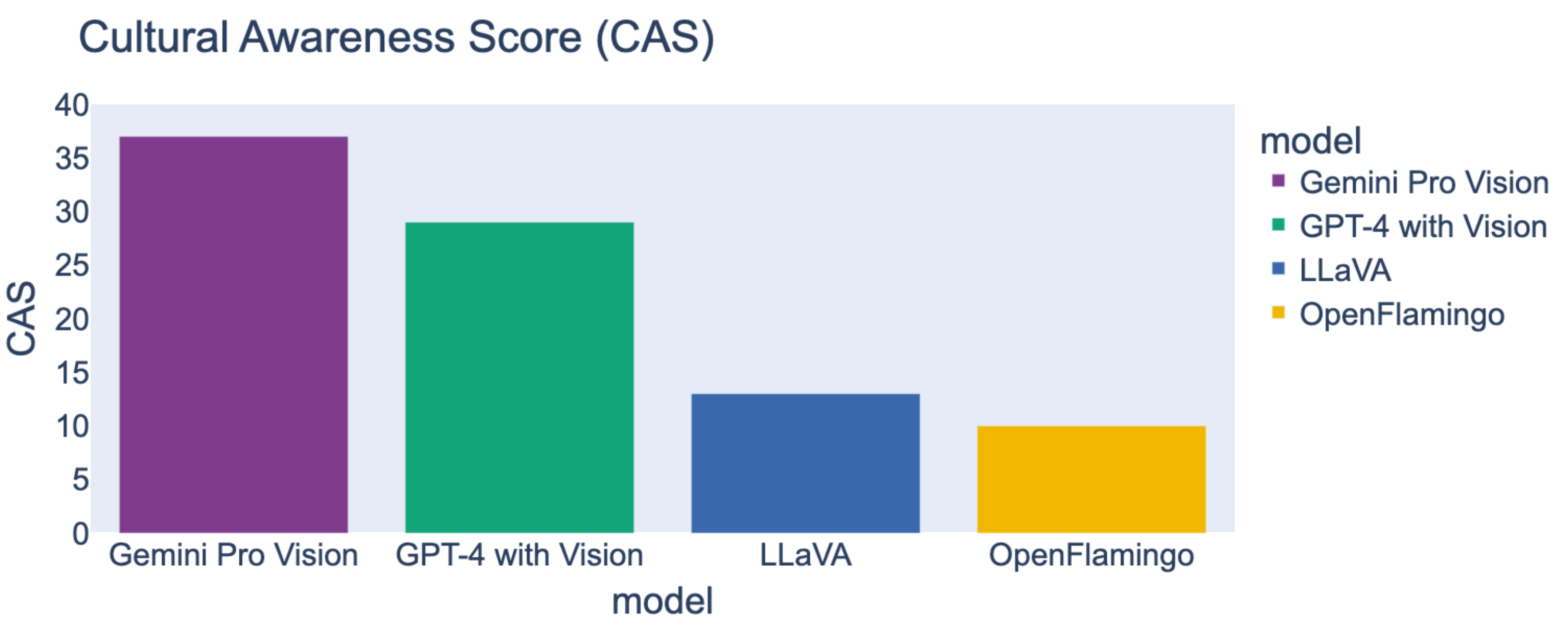}
        \caption{CAS across the models chosen for experimentation.}
        \label{fig:7}
\end{figure}

When dissecting the content of the images, a pattern emerged: real-life photographs, particularly those depicting dancers, not only achieved the highest CAS scores but also showed the lowest instances of erroneous image captioning. 
The analysis extended to illustrations and photographs that represented cultural concepts and mythologies. These categories were more prone to inaccuracies, leading to lower Cultural Awareness Scores. Moreover, vector images and icons denoting cultural signs and symbols were marked by the most significant levels of hallucinations, accompanied by the lowest Cultural Awareness Scores among the assessed categories.
The overall distribution of results for the Cultural Awareness Scores (CAS) for all models is presented in Fig. \ref{fig:7}.
A nuanced observation was made regarding the frequency of caption distortions, which depended on image type. An area of concern was identified in the performance of GPT-4V, where approximately 55\% of the image captions were deemed failures. This issue was particularly pronounced in the depiction of signs and symbols, pointing to a need for improvement in the accuracy of automated image captioning systems.

We can conclude that the CAS is low for all four selected vision-language models overall, and there is still significant room for improvement in image captioning of culture-specific concepts. The labeled dataset MOSAIC-1.5k, containing ground truth and assigned binary CAS can be used for model fine-tuning and generating Cultural Awareness Scores on unseen data. 
\section{Conclusion}
Captioning images related to mythology, folk dances, cultural signs, and symbols is a challenging task. Often, images are mislabeled or represented only as a description of the object in detail. Rich cultural background information and context are lost. We often observe confusion between similar or related cultural concepts in vision-language models. 

Our research aims to evaluate the performance of four vision-language models using the newly introduced Cultural Awareness Score, CAS. In addition to proposing and defining the CAS, we offer a labeled dataset MOSAIC-1.5k. This dataset contains 1,500 images of cultural concepts labeled with ground truth by humans, as well as predictions with assigned Cultural Awareness Scores that can be used for downstream evaluation of unseen image captions. We include several image examples for each cultural concept to reduce bias and increase accuracy. 

Further research is necessary to increase the initial dataset's size, to apply Cultural Awareness Scores to new data, and to evaluate other vision-language models.

\section{Ethical Statement}
In conducting our research, we have utilized a meticulously curated dataset featuring images from diverse cultural backgrounds while strictly adhering to ethical guidelines for data representation and use. 
We recognize the sensitivity surrounding cultural symbols and narratives, which hold deep significance for various communities. Our intention is to elevate the understanding and appreciation of cultural diversity through AI, mitigating the risk of cultural misrepresentation or stereotyping. We have carefully considered the benefits of employing these culturally rich images to advance our research against the potential risks of misinterpretation or misuse.

In alignment with principles of fairness and inclusivity, our work is dedicated to transparently sharing our research methodology, findings, and the considerations involved in developing the CAS and curating our dataset. This commitment extends to disclosing potential biases in our AI models and datasets, with ongoing efforts to address and mitigate these biases. Our approach aims to make the rationale behind our model evaluations and dataset curation clear, fostering an environment of ethical research practices that prioritize cultural understanding and sensitivity.


\vspace{12pt}

\begin{thebibliography}{00}
\bibitem{b1} J. Achiam et al., ``GPT-4 Technical Report,'' arXiv.org, Mar. 15, 2023. https://arxiv.org/abs/2303.08774
\bibitem{b2} J.-B. Alayrac et al., ``Flamingo: a Visual Language Model for Few-Shot Learning'' arXiv.org, Apr. 29, 2022. https://arxiv.org/abs/2204.14198
\bibitem{b3} R. Anil et al., ``Gemini: a family of highly capable multimodal models,'' arXiv.org, Dec. 19, 2023. https://arxiv.org/abs/2312.11805
\bibitem{b4} A. Awadalla et al., ``OpenFlamingo: an Open-Source framework for training large autoregressive Vision-Language models,'' arXiv.org, Aug. 02, 2023. https://arxiv.org/abs/2308.01390
\bibitem{b5} S. Bai et al., ``TouchStone: Evaluating Vision-Language Models by Language Models'' arXiv.org, Aug. 31, 2023. https://arxiv.org/abs/2308.16890
\bibitem{b6} O. Burda-Lassen, ``Machine Translation of Folktales: small-data-driven and LLM-based approaches,'' ACL Anthology, Sep. 01, 2023. https://aclanthology.org/2023.clasp-1.8/
\bibitem{b7} O. Burda-Lassen, ``Ukrainian-To-English Folktale Corpus: Parallel Corpus Creation and augmentation for Machine Translation in Low-Resource Languages,'' ACL Anthology, Sep. 01, 2022. https://aclanthology.org/2022.amta-coco4mt.4/
\bibitem{b8}Y. Cao, W. Li, J. Li, Y. Yuan, A. Karamolegkou, and D. Hershcovich, “Exploring Visual Culture Awareness in GPT-4V: A Comprehensive Probing,” arXiv.org, Feb. 08, 2024. https://arxiv.org/abs/2402.06015
\bibitem{b9} E. Cetinic, ``Iconographic image captioning for artworks,'' arXiv.org, Feb. 07, 2021. https://arxiv.org/abs/2102.03942
\bibitem{b10} W. Dai et al., ``InstructBLIP: Towards General-purpose Vision-Language Models with Instruction Tuning,'' arXiv.org, May 11, 2023. https://arxiv.org/abs/2305.06500
\bibitem{b11} C. Fu et al., ``MME: A Comprehensive Evaluation Benchmark for Multimodal Large Language models'' arXiv.org, Jun. 23, 2023. https://arxiv.org/abs/2306.13394
\bibitem{b12} S. Huang et al., ``Language Is Not All You Need: Aligning Perception with Language Models,'' arXiv.org, Feb. 27, 2023. 
\bibitem{b13}J. Kharchenko, T. Roosta, A. Chadha, and C. Shah, “How well do LLMs represent values across cultures? Empirical analysis of LLM responses based on Hofstede Cultural dimensions,” arXiv.org, Jun. 21, 2024. https://arxiv.org/abs/2406.14805https://arxiv.org/abs/2302.14045
\bibitem{b14} V. Krishna et al., ``IMAGINATOR: Pre-Trained Image+Text Joint Embeddings using Word-Level Grounding of Images,'' arXiv.org, May 12, 2023. https://arxiv.org/abs/2305.10438
\bibitem{b15} C. Li et al., ``Multimodal foundation models: from specialists to General-Purpose Assistants,'' arXiv.org, Sep. 18, 2023. https://arxiv.org/abs/2309.10020
\bibitem{b16} Y. Li, Y. Du, K. Zhou, J. Wang, W. X. Zhao, and J.-R. Wen, ``Evaluating object hallucination in large Vision-Language models,'' arXiv.org, May 17, 2023. https://arxiv.org/abs/2305.10355
\bibitem{b17} H. Liu, C. Li, Q. Wu, and Y. J. Lee, ``Visual instruction tuning,'' arXiv.org, Apr. 17, 2023. https://arxiv.org/abs/2304.08485
\bibitem{b18} B. McKinzie et al., ``MM1: Methods, Analysis \& Insights from Multimodal LLM Pre-training,'' arXiv.org, Mar. 14, 2024. https://arxiv.org/abs/2403.09611
\bibitem{b19}S. Nayak et al., “Benchmarking Vision language models for cultural understanding,” arXiv.org, Jul. 15, 2024. https://arxiv.org/abs/2407.10920
\bibitem{b20} S. Sheng, A. N. Venkitasubramanian, and M.-F. Moens, ``A Markov network based passage retrieval method for multimodal question answering in the cultural heritage domain,'' in Lecture notes in computer science, 2018, pp. 3–15.  
\bibitem{b21} S. M. T. I. Tonmoy et al., ``A comprehensive survey of hallucination mitigation techniques in large language models,'' arXiv.org, Jan. 02, 2024. https://arxiv.org/abs/2401.01313
\bibitem{b22} J. Wang et al., ``GIT: a generative image-to-text transformer for vision and language,'' arXiv.org, May 27, 2022. https://arxiv.org/abs/2205.14100
\bibitem{b23} S. Yin et al., ``Woodpecker: Hallucination Correction for Multimodal large Language models,'' arXiv.org, Oct. 24, 2023. https://arxiv.org/abs/2310.16045
\bibitem{b24} B. Zheng et al., ``Image captioning for cultural artworks: a case study on ceramics,'' Multimedia Systems, vol. 29, no. 6, pp. 3223–3243, Sep. 2023, doi: 10.1007/s00530-023-01178-8.
\end{thebibliography}
\end{document}